\documentclass[runningheads]{llncs}

\usepackage{eccv}

\usepackage{eccvabbrv}

\usepackage{graphicx}
\usepackage{booktabs}
\usepackage{amsmath}
\usepackage{amssymb}

\usepackage[accsupp]{axessibility}  % Improves PDF readability for those with disabilities.

\usepackage{hyperref}

\usepackage{orcidlink}

\begin{document}

% ---------------------------------------------------------------
\title{EgoCogNav: Cognition-aware Human Egocentric Navigation}

\titlerunning{EgoCogNav: Cognition-aware Human Egocentric Navigation}

\author{Zhiwen Qiu\inst{1} \and
Ziang Liu\inst{1} \and
Wenqian Niu\inst{2} \and
Tapomayukh Bhattacharjee\inst{1} \and
Saleh Kalantari\inst{1}}

\authorrunning{Z.~Qiu et al.}

\institute{
  Cornell University, Ithaca, NY, USA \and
  Georgia Institute of Technology, Atlanta, GA, USA
}

\maketitle

\begin{abstract}
Modeling the cognitive and experiential factors of human navigation is central to deepening our understanding of human–environment interaction and to enabling safe social navigation and effective assistive wayfinding. Most existing methods focus on forecasting motions in fully observed scenes and often neglect human factors that capture how people feel and respond to space. To address this gap, we propose EgoCogNav, a multimodal egocentric navigation framework that jointly forecasts perceived path uncertainty, trajectories and head motion from egocentric video, gaze, and motion history. To facilitate research in the field, we introduce the Cognition-aware Egocentric Navigation (CEN) dataset consisting of 6 hours real-world egocentric recordings capturing diverse navigation behaviors in real-world scenarios. Experiments show that EgoCogNav learns the perceived uncertainty that strongly correlates with human-like behaviors such as scanning, hesitation, and backtracking while improving trajectory and head-motion forecasting on held-out navigation recordings. \noindent Project page: \url{https://calvinzqiu.github.io/egocognav-project/}
\end{abstract}
\section{Introduction}
\label{sec:intro}

\begin{figure*}[t]
  \centering
  \includegraphics[width=1\linewidth]{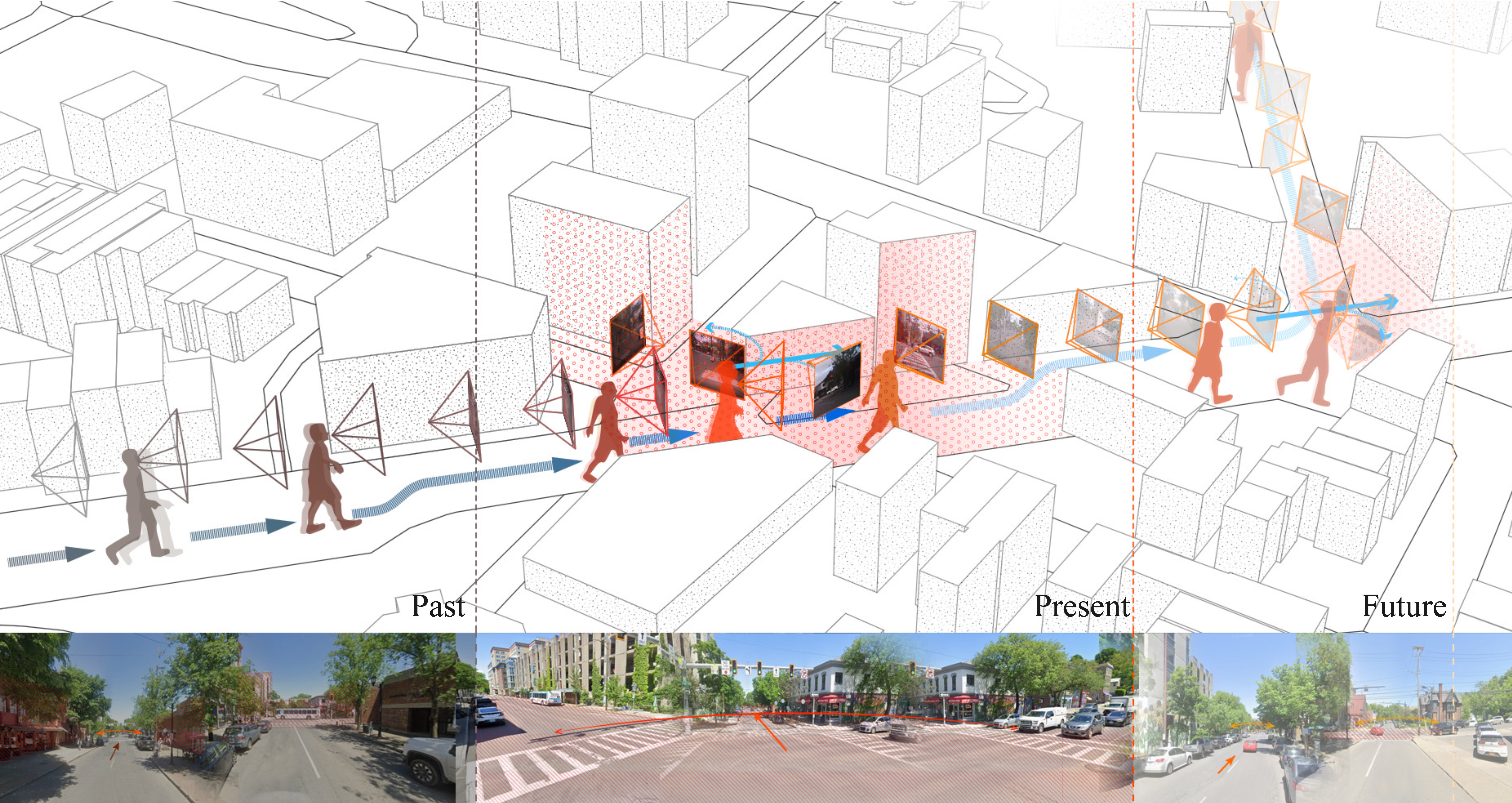}
  \caption{Given a past window of motion, head rotations, gaze, and navigational goal, our model jointly predicts a future body-frame trajectory, head poses, and the current state of perceived uncertainty. This setting reflects real-world navigation challenges in environments where the model must anticipate internal cognitive state and making decisions for subsequent head motion and movement.}
  \label{fig:problem}
\end{figure*}

Understanding human navigation processes is fundamental to safe, reliable human-environment and human-machine interactions \cite{farr2012wayfinding, raubal1999formal}. Humans perceive their environments from egocentric perspectives to inform decision-making for subsequent motions. Learning such cues is crucial for applications including autonomous driving \cite{jain2020discrete, vellenga2024evaluation}, social robot navigation \cite{salzmann2023robots, mavrogiannis2022social, walker2022influencing}, and assistive wayfinding systems \cite{yang2025path, edward2009cognitive, pan2006computational}. 

Modeling human navigation requires not only accurate scene representation and motion forecasting, but also human factors that capture how people experience and emotionally respond to built environments, which are closely related to productivity, satisfaction, and well-being \cite{raubal2001human, dubey2019fusion, kaya2016importance, mackett2021mental}. Incorporating these cognitive states into behavioral models can deepen our understanding of interacting with environments and substantially enhance their utility for salient environmental design and personalized navigation assistance in complex settings \cite{kaya2016importance, devlin2014wayfinding}. Perceived uncertainty, defined as a state in which an individual is trying to decide between alternative courses of action, is a driver of wayfinding behaviors tied to discomfort and negative emotions \cite{hirsh2012psychological, devlin2014wayfinding, pouyan2024elderly}. Predicted uncertainty levels can inform probabilistic wayfinding behaviors as well as a valuable metric for understanding users' experience of a space. However, most prior work in human trajectory prediction does not account for cognitive and psychological processes and relies instead on motion history and context for path forecasting \cite{jeong2025multi, mao2023leapfrog, gu2022stochastic}, use rule- or agent-based methods \cite{maruyama2017simulation, zhu2021follow, de2023large}, or develop socially compliant policies for navigation \cite{kretzschmar2016socially, hirose2023sacson, nguyen2023toward, song2024vlm}. Although some incorporate cognitive factors such as emotions, panic, and stress, these frameworks typically assume fully observed third-person or bird’s-eye-view (BEV) scenes and often planar worlds \cite{edward2009cognitive, bosse2013modelling, pan2006computational, yang2025path}. Furthermore, there are limited multimodal egocentric navigation datasets, especially those with cognitive annotations, to study these effects at scale.

To address the challenges, we propose EgoCogNav, a multimodal egocentric navigation framework that jointly predicts perceived uncertainty, body-centered trajectories and head motion. The framework fuses first-person scene evidence from a pre-trained DINOv2 vision backbone \cite{caron2021emerging, oquab2023dinov2} with recent motion where cognition and behaviors are learned in a single perception–decision–action loop. The design is modular and extensible for additional modality input with shared time-series forecasting module. Finally, we introduce the Cognition-Aware Egocentric Navigation (CEN) dataset that features rich multimodal streams from 17 participants across diverse indoor and outdoor scenes for human navigation research. Our contribution can be summarized as follows: (1) We formalize cognition-aware egocentric forecasting task to jointly predict trajectory, head motion, and moment-to-moment perceived path uncertainty; (2) We propose EgoCogNav, which fuses multiple sensory inputs with human-grounded uncertainty to produce behaviorally realistic behavior forecasts useful for assistive navigation; (3) We introduce a dataset consisting of 6 hours of real-world human navigation sessions that covers diverse sites and environmental conditions.

\section{Related Work}
\label{sec:relatedwork}

\noindent\textbf{Human trajectory prediction.} The prospect of predicting pedestrian trajectories from third-person or BEV has been extensively studied. Prior methods include deterministic predictors that estimate one most likely path using scene information and social cues \cite{alahi2016social, kothari2021human, saadatnejad2023social}, and stochastic models that generate multimodal futures or full distributions from probabilistic frameworks such as diffusion models \cite{bae2024singulartrajectory, gu2022stochastic, mao2023leapfrog} and normalizing flows \cite{maeda2023fast,scholler2021flomo}. However, third-person views often neglect first-person natural perceptual and cognitive signals, which can limit the development of high-fidelity modeling of human motion in cluttered environments \cite{mavrogiannis2023core}. Despite substantial progress in egocentric human motion estimation \cite{li2023ego, tome2019xr, yi2025estimating, akada20243d}, trajectory and motion forecasting that incorporate human cognitive states remain underexplored. Wang and colleagues \cite{wang2024egonav} use a diffusion framework to generate multiple plausible future trajectories from RGB-D video with a learned visual-memory of the surroundings. Pan and colleagues \cite{pan2025lookout} predict 6-DoF head poses to learn collision-aware, information-seeking behaviors by projecting 3D-feature volumes into BEV. However, these approaches largely assume homogeneous behaviors and overlook individual differences and internal cognitive states that are salient in real-world navigation. Integrating experiential factors can reveal why and when people hesitate or backtrack to improve behavioral fidelity and enable assistance and design interventions that anticipate difficulties rather than merely reacting to observed motion.

\noindent\textbf{Perceived uncertainty in human wayfinding models.} Perceived path uncertainty is a key cognitive variable closely linked to wayfinding performance which emerges from limited knowledge about forthcoming events \cite{devlin2014wayfinding, pouyan2024elderly}. The Entropy Model of Uncertainty (EMU) \cite{hirsh2012psychological} posits that uncertainty is affected by the range of perceived choices and peaks when perceived probability for each choice is equal. Prior strategies to integrate human cognition typically include translating empirical observations like "go-to" affordances into rule-based deterministic models \cite{raubal2001human}, or defining utility parameters such as path choice and individual preference for agents to optimize during navigation \cite{zhu2023behavioral, xie2022simulation, huang2023modeling}. For instance, Huang and colleagues \cite{huang2023modeling} present a two-layer floor-field cellular automaton with three intertwined sub-modules (exit choice, locomotion, and exit-choice switching) to model risk- and uncertainty-aware decisions to capture pedestrian dynamics. Yang and colleagues \cite{yang2025path} introduce a probabilistic data-driven wayfinding agent that integrates isovist-derived spatial metrics, signage visibility/recency, route-choice counts, and individual factors to predict continuous perceived path uncertainty for simulating human trajectories in indoor settings. In contrast to prior BEV- or rule-based formulations, we employ a data-driven and learning-based method that predicts perceived path uncertainty from multimodal first-person cues.

\noindent\textbf{Memory-augmented motion prediction.} Recent works have explored memory mechanisms to extend the effective context of motion prediction models. Xu and colleagues \cite{xu2022remember} introduce an instance memory bank (MemoNet) that stores and retrieves past trajectory patterns to improve pedestrian forecasting. Shi and colleagues \cite{shi2024mtr++} introduces MTR++ that utilizes learnable intention queries that capture recurring motion patterns across agents that provides a form of parametric memory that the decoder attends to during prediction. Similarly, Yu and colleagues \cite{yu2024fmtp} use learnable trajectory anchors as reference patterns that encode frequently observed movement modes, enabling the decoder to attend to relevant anchors and produce diverse yet plausible future trajectories. In the generative domain, Zhang and colleagues \cite{zhang2023remodiffuse} augment diffusion-based motion generation (ReMoDiffuse) with a retrieval mechanism that queries a database of motion clips. While these approaches demonstrate the value of memory for motion forecasting, few focus on conditioning memory retrieval or processing on predicted human cognition state.

\noindent\textbf{Egocentric dataset for human navigation.} Many egocentric datasets \cite{damen2018scaling,grauman2022ego4d,li2021ego,lv2024aria} provide real-world monocular RGB videos of diverse daily activities such as household, workplace, and outdoor scenes. To enrich human sensory inputs, subsequent efforts were made to incorporate wearable and instrumented platforms like motion-capture suits and Project Aria glasses \cite{engel2023project} to deliver calibrated multimodal signals, including head poses \cite{wang2023scene,pan2025lookout}, gaze \cite{lv2024aria,li2021eye}, hand and object tracking \cite{banerjee2025hot3d,tang2023egotracks}, full-body motion \cite{ma2024nymeria}, and detailed scene and action annotations \cite{li2024egoexo}.  However, these datasets are not explicitly curated for human navigation. Few works utilized synthetic pipelines to simulate virtual human navigation and collisions in 3D environments from multi-view body-mounted cameras, but resulting motions are often simplified or unnatural due to constraints in existing motion generation methods. Closer to our setting, Wang and colleagues \cite{wang2024egonav} and Pan and colleagues \cite{pan2025lookout} develop egocentric human-navigation datasets, yet the coverage is comparatively limited and the focus is primarily trajectory or head-pose forecasting without leveraging varied sensory inputs or human cognitive factors. Moreover, neither dataset is publicly released to date. Therefore, we introduce a new dataset that unifies sensory signals, cognitive indicators, and accurate localization across diverse indoor and outdoor scenes.

\section{Method}
\label{sec:method}

\begin{figure*}[t]
\centering
\includegraphics[width=1\linewidth]{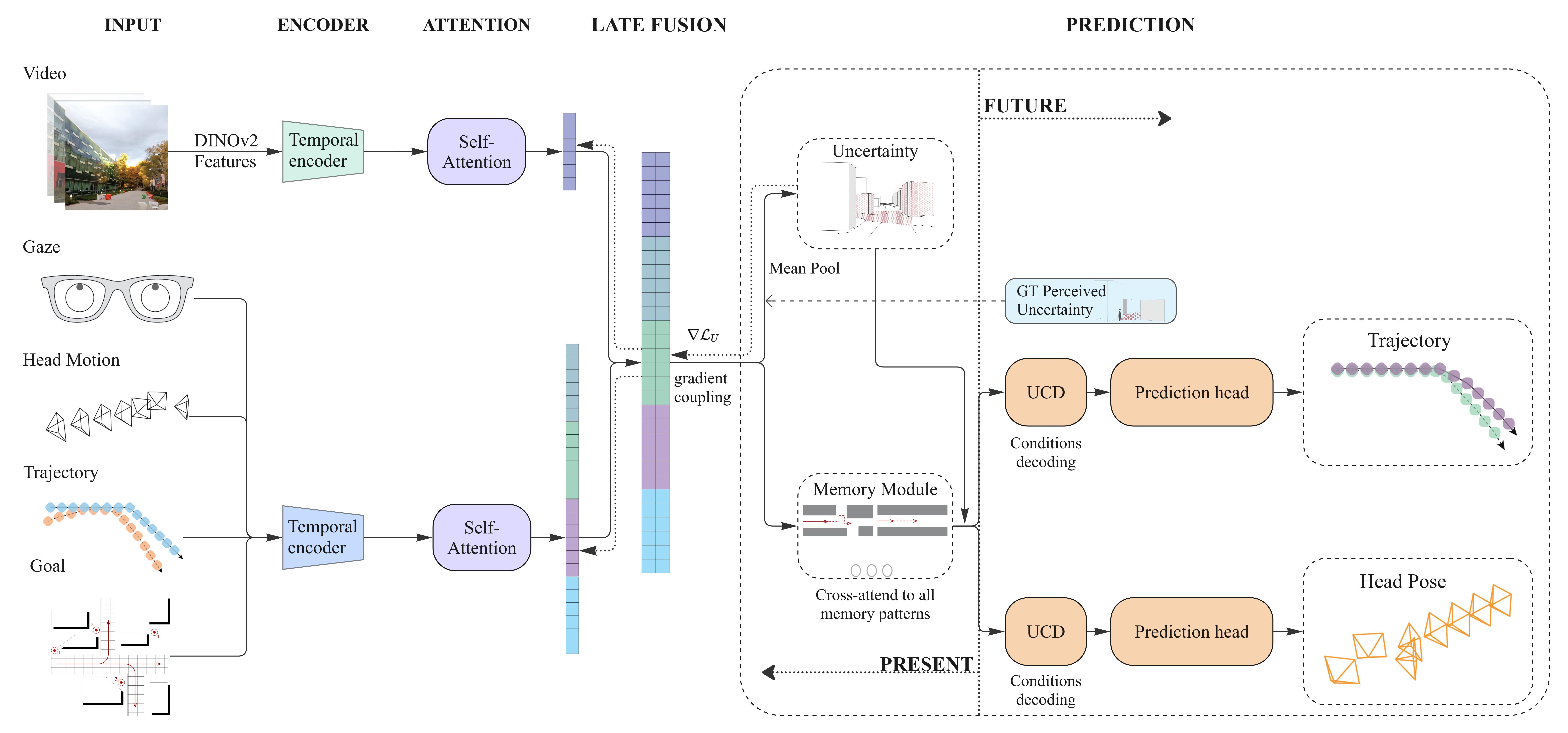}
\caption{\textbf{EgoCogNav architecture.} Given past egocentric video and sensor inputs, the model encodes perception (video features from a pre-trained vision transformer) and action (past motion, gaze, and goal) into two self-attention streams and fuses them by late concatenation. A cognition module predicts perceived uncertainty and retrieves situation-relevant context from learnable memory patterns, and then conditions decoding via adaptive layer normalization to forecast future trajectory and head motion.}
\label{fig:architecture}
\end{figure*}

\subsection{Problem Formulation}
As illustrated in Fig.~\ref{fig:problem}, given a past egocentric video $\mathbf{X}_{1:T_1}\!\in\!\mathbb{R}^{T_1\times H\times W\times 3}$, past body-frame motion $\mathbf{S}_{1:T_1}=[\Delta x,\Delta y,\Delta\psi]\!\in\!\mathbb{R}^{T_1\times 3}$, 6D continuous head rotations~\cite{zhou2019continuity} $\mathbf{H}_{1:T_1}\!\in\!\mathbb{R}^{T_1\times 6}$, normalized gaze points $\mathbf{G}_{1:T_1}\!\in\!\mathbb{R}^{T_1\times 2}$, and body-frame navigation goal $\mathbf{q}\!=\![d,\sin\beta,\cos\beta]\!\in\!\mathbb{R}^{3}$ where $d$ is distance and $\beta$ the goal bearing, we jointly predict the future trajectory $\widehat{\mathbf{S}}_{T_1\!+\!1:T_1\!+\!T_2}\!\in\!\mathbb{R}^{T_2\times 3}$, head-pose sequence $\widehat{\mathbf{H}}_{T_1\!+\!1:T_1\!+\!T_2}\!\in\!\mathbb{R}^{T_2\times 6}$, and the current perceived uncertainty $\hat{U}_{T_1}\!\in\![0,1]$. At 10\,Hz, we use a past window $T_1\!=\!30$ steps (3\,s) and future horizon $T_2\!=\!10$ steps (1\,s) which matches the timescale of short-term navigation behaviors such as hesitation and rapid head reorientation \cite{keller2020uncertainty, brunye2018spatial}.

\subsection{EgoCogNav Architecture}
Our framework is organized into three modules (Fig.~\ref{fig:architecture}): (1)~a perception module that extracts spatio-temporal features from recent video using a pre-trained vision transformer, (2)~an action module that encodes past body-frame motion, head rotation, and gaze together with goal conditioning, and (3)~a cognition module that predicts the current perceived uncertainty $\hat{U}_t$ and uses it to condition decoding via adaptive layer normalization and to augment features with learnable memory modules. The perception and action streams are encoded independently with self-attention and then fused through late concatenation into a shared representation from which we forecast body-frame trajectory and head motion, together with a cognition head that estimates perceived uncertainty.

\noindent\textbf{Perception module.} The perception module processes the past RGB frames. Each frame is resized to $224{\times}224$ and passed through a frozen pre-trained DINOv2~\cite{caron2021emerging,oquab2023dinov2} vision transformer. We use the per-frame \texttt{CLS} token as a global visual descriptor to produce a temporal stack of video features $\mathbf{F}^{\text{vid}}\in\mathbb{R}^{T_1\times 384}$. These features are linearly projected to a shared model dimension $d$ and fed into the subsequent fusion module.

\noindent\textbf{Action module.} We encode three synchronized cues over the past $T_1$ steps: body-frame trajectory deltas $\mathbf{S}_{1:T_1}\!=\![\Delta x,\Delta y,\Delta\psi]$, head rotations $\mathbf{H}_{1:T_1}$, and gaze points $\mathbf{G}_{1:T_1}\!=\!(u,v)$ in normalized image coordinates, together with the navigation goal $\mathbf{q}\!=\![d,\sin\beta,\cos\beta]$ expressed in the current body frame. All streams are aligned with sinusoidal positional encoding and projected to the same width $d$ before fusion.

\noindent\textbf{Multi-modal fusion.} We process the action and perception streams independently before fusion. The action stream concatenates body-frame motion, head rotation, gaze, and goal features into a single temporal sequence projecting to width $d$, and encodes it with a 4-layer transformer encoder using sinusoidal positional encoding. The perception stream similarly projects and encodes the DINOv2 feature sequence with a 2-layer transformer encoder. Both streams are temporally mean-pooled and concatenated to produce a fused representation $\mathbf{f}\!\in\!\mathbb{R}^{2d}$, which is then projected to $\mathbf{h}\!\in\!\mathbb{R}^{d}$ via a linear layer with layer normalization. This late fusion strategy lets each stream learn modality-specific temporal patterns through self-attention before combining them into a shared representation.

\noindent\textbf{Cognition module.} The cognition module is the core module of our architecture. It captures the internal cognitive state of the navigator and uses it to guide both how and what information the prediction heads use. It consists of three sub-components: (1)~gradient-coupled uncertainty prediction, (2)~memory-augmented prediction, and (3)~uncertainty-conditioned decoding (UCD) as described below.

\emph{Gradient-coupled uncertainty estimation.}
Conceptually aligned with the EMU theory of competing scene interpretations and action choices~\cite{hirsh2012psychological}, we model the navigator's current internal state as a single-step prediction of perceived uncertainty $\hat{U}_t \in [0,1]$, a state signal that summarizes moment-to-moment choice difficulty preceding navigation behaviors. The cognition head operates on the projected fused representation $\mathbf{h} \in \mathbb{R}^{d}$ with a two-layer MLP and sigmoid activation:
\begin{equation}
\label{eq:u_hat}
\hat{U}_t = \sigma\!\big(\text{MLP}(\mathbf{h})\big) \in [0,1].
\end{equation}
Since $\hat{U}_t$ is predicted from shared encoder features, the task objectives jointly shape the encoder representation to encourage features that support both motion prediction and uncertainty estimation.

\emph{Memory-augmented prediction.}
While predicting $\hat{U}_t$ provides useful gradient signal to the encoder, it does not extend what information is available to the forecasting heads. Navigation under perceived uncertainty often requires context beyond the immediate $T_1$-step input window with patterns from similar past situations. Inspired by learnable intention queries in motion prediction~\cite{shi2024mtr++}, we augment the model with $N_m = 16$ learnable navigation pattern vectors $\mathbf{M} \in \mathbb{R}^{N_m \times d_m}$ that capture recurring navigation situations from the training data. The current navigation state $\mathbf{h}$ queries these patterns via cross-attention to retrieve situation-relevant context:
\begin{equation}
\label{eq:memory}
\mathbf{c} = \text{CrossAttn}\!\big(W_q \!\cdot\! \mathbf{h},\;
\mathbf{M},\;\mathbf{M}\big)
\end{equation}
\begin{equation}
\label{eq:memory_inject}
 \mathbf{h}^{\text{mem}} = \mathbf{h} + W_{\text{out}} \!\cdot\! \mathbf{c}
\end{equation}
where $W_q$ projects to the pattern space and $W_{\text{out}}$ projects back, with $W_{\text{out}}$ zero-initialized so the module starts as identity and gradually learns to contribute.

\emph{Uncertainty-conditioned decoding.}
While memory extends what information is available, it does not inform the prediction heads about the current level of perceived uncertainty of the navigator. To this end, we introduce UCD that modulates the shared latent representation based on the predicted cognitive cost. We utilize the adaptive layer normalization from~\cite{peebles2023scalable} to condition on the predicted $\hat{U}_t$. Since $\hat{U}_t$ is predicted from the shared encoder rather than provided as an external input, UCD treats uncertainty as a learned internal state and lets the model learn how to map it into modulation parameters. Given the memory-augmented features $\mathbf{h}^{\text{mem}} \in \mathbb{R}^{d}$, UCD produces uncertainty-aware features via:
\begin{equation}
\label{eq:ucd}
\tilde{\mathbf{h}} = (1 + \gamma) \odot
\text{LN}(\mathbf{h}^{\text{mem}}) + \beta, \quad
 (\gamma, \beta) = \text{MLP}(\hat{U}_t)
\end{equation}
where $\text{LN}$ is layer normalization, $\odot$ is element-wise multiplication, and the MLP maps $\hat{U}_t \in \mathbb{R}$ to $\mathbb{R}^{2d}$ via a two-layer network with SiLU activation~\cite{elfwing2018sigmoid}. Following~\cite{peebles2023scalable}, the MLP is zero-initialized so that $\gamma = \mathbf{0}$ and $\beta = \mathbf{0}$ at the start of training to ensure identity behavior until the model learns meaningful modulation.

Memory and UCD address complementary limitations. Memory extends the information available with learned navigation patterns, and UCD adjusts how the prediction heads internally process this information based on perceived uncertainty of the current scene. 

\noindent\textbf{Trajectory and head motion prediction.} 
On top of the uncertainty-aware features $\tilde{\mathbf{h}}$, two prediction heads produce task-specific forecasts: a 3-DOF body-frame trajectory $\widehat{\mathbf{S}}_{T_1+1:T_1+T_2}$ and a 6D head-motion sequence $\widehat{\mathbf{H}}_{T_1+1:T_1+T_2}$.

\subsection{Training Objectives}
\label{sec:loss}
\noindent\textbf{Task losses.}
For trajectory, we prioritize near-future steps~\cite{amit2020discount} with discounted $\ell_1$ loss and a variance regularization term:
\begin{equation}
\label{eq:traj_loss}
\mathcal{L}_{\text{traj}} = \textstyle\sum_{i=1}^{T_2}\gamma^{i}\,\big\|\widehat{\mathbf{S}}_i -
\mathbf{S}^{\text{gt}}_i\big\|_{1} + \lambda_{\text{var}}\big\|\operatorname{std}_{i}(\widehat{\mathbf{S}}) -
\operatorname{std}_{i}(\mathbf{S}^{\text{gt}})\big\|_{2}^{2}
\end{equation}
where $\gamma\!=\!0.98$ and $\lambda_{\text{var}}\!=\!0.3$, and $\boldsymbol{\sigma}(\cdot)\!\in\!\mathbb{R}^{3}$ is the per-coordinate standard deviation over the $T_2$ future steps. Following~\cite{pan2025lookout}, head rotations use the rotation matrix $\ell_1$ distance:
\begin{equation}
\label{eq:head_loss}
\mathcal{L}_{\text{head}}
= \frac{1}{T_2}\sum_{i=1}^{T_2}\big\|\widehat{\mathbf{R}}_{t+i}\,\mathbf{R}_{t+i}^{\top}-\mathbf{I}\big\|_{1}
\end{equation}
where $\widehat{\mathbf{R}}_{t+i}$ and $\mathbf{R}_{t+i}$ are rotation matrices recovered from the predicted and ground-truth 6D representations. We regress the human self-reported perceived uncertainty with mean squared error:
\begin{equation}
\label{eq:u_loss}
\mathcal{L}_{U}=\big\|\hat{U}_t-U^{\mathrm{human}}_t\big\|_{2}^{2}.
\end{equation}

\noindent\textbf{Multi-task objective.}
The total training loss combines all three task losses with equal weighting:
\begin{equation}
\label{eq:total_loss}
\mathcal{L} = \mathcal{L}_{\text{traj}} + \mathcal{L}_{\text{head}} + \mathcal{L}_{U}
\end{equation}

\subsection{Implementation Details}
We train on a single NVIDIA RTX 4090 with AdamW~\cite{loshchilov2017decoupled}, using cosine annealing over 300 epochs with a maximum learning rate of $1{\times}10^{-4}$, weight decay $8{\times}10^{-5}$, and batch size 64. DINOv2 features are precomputed and cached and the vision backbone remains frozen throughout training. The model dimension is $d\!=\!512$. The action encoder uses 4 transformer layers and the perception encoder uses 2 layers.  The memory module contains $N_m\!=\!16$ slots with internal dimension $d_m\!=\!256$.

\section{Cognition-aware Egocentric Navigation (CEN) Dataset}
\label{sec:dataset}

As discussed in Section~\ref{sec:relatedwork}, there is currently no publicly available dataset to support research into egocentric human navigation with cognitive factors. We therefore collect a multimodal egocentric navigation dataset combining rich sensory inputs with moment-to-moment annotations of perceived path uncertainty. We detail the data-collection pipeline and dataset statistics below.

\subsection{Data Collection}
\noindent\textbf{Hardware.} To accommodate the distinct characteristics of indoor and outdoor environments, we employ two complementary setups. For outdoors, we use Tobii Pro Glasses capturing 20fps RGB with high-quality binocular gaze and 6-axis IMU commonly used across studies \cite{onkhar2024evaluating}, and a reliable Garmin handheld GPS providing precise global positions. For indoor scenarios, we use Project Aria glasses \cite{engel2023project} that capture 20fps RGB video with accurate SLAM via two monochrome cameras, eye-tracking, and an IMU sensor. In both settings, participants hold an Xbox controller \footnote{https://xboxdesignlab.xbox.com/en-us/controllers/xbox-wireless-controller} to continuously self-report perceived path uncertainty on a normalized [0,1] scale.

\noindent\textbf{Recording procedure.} The procedure was approved by the Institutional Review Board (IRB). Before each session, participants were briefed on the study goals and received a tutorial on how to continuously report perceived uncertainty with the joystick for full range. Participants were explicitly instructed to perform route-finding behaviors when uncertain such as scanning the surroundings for cues, confirming signage or landmarks, and were reminded to check for passing vehicles before crossing streets. During recording, participants navigated from a fixed starting point through an identical sequence of waypoints per scenario toward predefined goals for each scenario. A researcher trailed each participant to present the next waypoint image upon arrival at previous one to ensure proper task progression.

\noindent\textbf{Data processing.} Outdoor recordings from Tobii Pro Glasses are processed in Tobii Pro Lab\footnote{https://www.tobii.com/products/software/behavior-research-software/tobii-pro-lab} and exported as \texttt{.tsv} files with multimodal streams including 2D gaze coordinates, 6-axis IMU signals, and egocentric videos. Trajectories are obtained from GPS logs in \texttt{.gpx} files and are smoothed with a Savitzky–Golay filter \cite{schafer2011savitzky}. Joystick signals are saved as \texttt{.csv} with magnitude of each press. Indoor sessions are stored in \texttt{.vrs} files and processed with Aria Machine Perception Services\footnote{https://huggingface.co/projectaria} to extract RGB video, 6D head-pose trajectories, scene point clouds, and eye-gaze estimates. All recordings are synchronized to a 10 Hz timeline via down-sampling with nearest neighbor or interpolation. We additionally annotate videos with behavior types and environment categories to provide labels for supervised learning and validation.

\noindent\textbf{Privacy.} All recordings were de-identified by blurring faces and removing audio to protect privacy of participants. 

\subsection{Dataset Statistics.} The dataset comprises approximately 6 hours from 17 participants with total 226k RGB frames across 42 distinct sites. Diverse sites were selected across indoor and outdoor settings, including university campuses, healthcare facilities, urban commercial streets, and natural routes. The sites are recorded at varied times of day with varying lighting, crowding, and traffic. Wayfinding routes are designed to integrate four uncertainty-inducing spatial types. We further annotate three label groups for stratified analysis and auxiliary supervision. Labels were selected from pilot observations and theory that cause choice difficulty and subsequent behavioral responses. Environment types include multiple-route junctions (\texttt{JCT}), occluded/poor-signage segments (\texttt{OCC}), dynamic/crowded areas (\texttt{CROWD}), and spatial transitions/sudden changes (\texttt{ST}); trajectory behaviors include hesitation/pausing (\texttt{HES}), wrong turn (\texttt{WRONG}), and backtrack (\texttt{BACK}); head-movement behaviors include information gathering (\texttt{SCAN}), information confirmation (\texttt{CONFIRM}), and look-back (\texttt{LB}). 

The four environment types account for 42.9\% of the recorded traversal, including junctions (\texttt{JCT}, 24.4\%), occluded or poor-signage areas (\texttt{OCC}, 10.6\%), spatial transitions or sudden scene changes (\texttt{ST}, 3.4\%), and crowded or dynamic areas (\texttt{CROWD}, 4.5\%). The remaining 57.1\% is treated as neutral. Mean self-reported uncertainty is substantially higher in navigation-challenging regions, reaching 0.47 at junctions and 0.44 in occluded or poorly signed areas, compared with 0.19 in neutral segments. This distribution shows that the joystick signal is not uniform across the route, but systematically increases in spatial conditions that require route choice, visual search, or additional confirmation. We further validate that the uncertainty signal reflects perceived navigation difficulty by comparing it against independently video-annotated behaviors labeled by two independent reviewers. Reported uncertainty is consistently elevated during decision-difficulty behaviors, with medium-to-large effect sizes for wrong turns (Cohen's $d=0.80$), look-backs ($d=0.66$), and backtracking ($d=0.55$). 

\section{Experiments}
\label{sec:experiments}
We evaluate EgoCogNav with baselines and ablate key design choices in Section~\ref{sec:ablation}. We also provide qualitative analyses that visualize trajectory and head-pose forecasts and uncertainty estimation in Section~\ref{sec:qual}. Finally, we discuss limitations and future directions in Section~\ref{sec:limitation}. All results are reported on a held-out test split. Since participants follow the same waypoint sequences within each scenario, this split evaluates generalization to held-out traversal instances, head and gaze movements, and local egocentric observations.

\subsection{Quantitative Evaluation}
\label{sec:quant}
\noindent\textbf{Metrics.} For trajectory, we report Average Displacement Error (ADE), which captures the mean distance between predicted and actual positions at each timestep; and Final Displacement Error (FDE), which measures the distance at the trajectory endpoint. We evaluate the L1 loss \cite{pan2025lookout} used during training for head rotations. As for uncertainty, we report mean absolute error (MAE), Spearman rank correlation coefficient \cite{hauke2011comparison} over all scenarios and on top-20\% high-uncertainty subset. $\Delta$U measures the mean elevation of predicted uncertainty during annotated navigation behaviors relative to neutral segments to capture the model's sensitivity to behavioral difficulty.

\noindent\textbf{Baselines.}
Since this is a novel task, there are limited baselines to the best of our knowledge. The closest related works \cite{wang2024egonav, qiu2022egocentric, pan2025lookout} are pre-printed or have no publicly released code. As a result, we compare against the following baselines: (1) Constant Velocity (\texttt{Const\_Vel}) \cite{mavrogiannis2022social}, which extrapolates future body-frame translation from the last linear velocity and future head rotation from the last angular velocity, (2) Linear Extrapolation (\texttt{Lin\_Ext}) \cite{pan2025lookout}, which fits a per-axis linear model to the past translation and rotation sequences and projects them into the future, (3) a standard Multimodal Transformer (\texttt{M\_Transformer}) baseline that uses the same inputs and training protocol as our model but performs early fusion by concatenating embeddings with a single temporal transformer decoder, followed by linear heads for the three comprehensive tasks, and (4) EgoCast \cite{escobar2025egocast} where we adapt its forecasting module which was originally designed for full-body pose to our perceived uncertainty, trajectory, and head-motion prediction settings with additional layers, using the same frozen DINOv2 features as EgoCogNav to isolate the forecasting module from the visual backbone.

As for perceived uncertainty prediction, we compare with two baselines: (1) an EMU-entropy (\texttt{EMU}) theory proxy, which computes two signals of perceptual ambiguity from visual scenes and behavioral variability from short-horizon motion, and learns a linear combination to the human uncertainty label; and (2) a PATH-U-adapted (\texttt{PATH\_U}) \cite{yang2025path} model, as it was originally designed for indoor wayfinding using signage, we adapt it to the egocentric setting by composing a 5-dimensional feature vectors capturing decision complexity and behavioral variability (e.g., junction count, occlusion/poor-signage count, goal distance), and fitting a linear regressor to predict perceived uncertainty.

\begin{table}[t]
\centering
\small
\setlength{\tabcolsep}{3pt}
\renewcommand{\arraystretch}{1.15}
\begin{tabular}{@{}l|ccc|ccc|cc@{}}
\toprule
\textbf{Method} &
\multicolumn{3}{c|}{All Scenarios} &
\multicolumn{3}{c|}{High-U Scenarios} &
\multicolumn{2}{c}{Uncertainty} \\
\cmidrule(lr){2-4}\cmidrule(lr){5-7}\cmidrule(lr){8-9}
 & ADE $\downarrow$ & FDE $\downarrow$ & Head $\downarrow$
 & ADE $\downarrow$ & FDE $\downarrow$ & Head $\downarrow$
 & MAE $\downarrow$ & $\rho$ $\uparrow$ \\
\midrule
Const\_Vel     & .1892 & .4257 & .0875 & .2170    & .4778    & .0877    & --    & --   \\
Lin\_Ext       & -- & -- & .1224 & --    & --    & .1235    & --    & --   \\
M\_Transformer     & .1536 & .3213 & .0776 & .1684 & .3469 & .0778 & .1247 & .683 \\
EgoCast$^*$~\cite{escobar2025egocast}
 & .1092 & .2184 & .0712 & .1198 & .2369 & .0716 & .1029 & .752 \\
\textbf{EgoCogNav}
 & \textbf{.1051} & \textbf{.2074} & \textbf{.0698} & \textbf{.1155} & \textbf{.2256} & \textbf{.0704} &
\textbf{.0986} & \textbf{.788} \\
\bottomrule
\end{tabular}
\caption{\textbf{Baseline comparison.} Trajectory (ADE/FDE), head rotation (L1), and uncertainty value error (MAE) on the full test set and on a high-uncertainty subset.}
\label{tab:baseline}
\end{table}

\begin{table}[t]
\centering
\small
\setlength{\tabcolsep}{8pt}
\begin{tabular}{l|cc|cc|c}
\toprule
\textbf{Method} &
\multicolumn{2}{c|}{All Scenarios} &
\multicolumn{2}{c|}{High-U Scenarios} &
\\
\cmidrule(lr){2-3}\cmidrule(lr){4-5}
 & MAE $\downarrow$ & $\rho$ $\uparrow$ &
 MAE $\downarrow$ & $\rho$ $\uparrow$ &
$\Delta$U $\uparrow$ \\
\midrule
EMU \cite{hirsh2012psychological}   & .1887 & .081 & .1842 & .100 & .0122 \\
 PATH\_U~\cite{yang2025path}         & .1857 & .195 & .1814 & .210 & .0212 \\
\textbf{EgoCogNav}                  & \textbf{.0986} & \textbf{.788} & \textbf{.1017} & \textbf{.636} &
\textbf{.0829} \\
\bottomrule
\end{tabular}
\caption{\textbf{Uncertainty prediction.} MAE and Spearman $\rho$ are reported for all samples and for the high-uncertainty subset. $\Delta$U measures mean predicted uncertainty elevation during annotated navigation behaviors relative to neutral segments.}
\label{tab:uncertainty}
\end{table}

\noindent\textbf{Results.}
The comparison results are presented in Table~\ref{tab:baseline}. Our model achieves the best trajectory and head-motion performance on both the full test set and the high-uncertainty subset, reducing ADE/FDE by 3.8\%/5.0\% relative to the EgoCast-adapted baseline~\cite{escobar2025egocast}. The M\_Transformer baseline uses early fusion without cognition-aware modules and shows degraded performance, which suggests that late fusion with modality-specific temporal encoding better preserves the sensory patterns needed for accurate forecasting. Table~\ref{tab:uncertainty} directly compares our learned uncertainty against dedicated baselines to evaluate whether the model captures meaningful cognitive states. The EMU entropy proxy \cite{hirsh2012psychological} and PATH\_U heuristic \cite{yang2025path} both achieve near-chance rank correlation, indicating that hand-crafted rules based on scene ambiguity or decision complexity cannot capture the subjective, person-specific nature of perceived uncertainty. In contrast, our model achieves $\rho = 0.788$ by learning to map multimodal sensory-motor patterns to individual cognitive states through joint training. The behavioral sensitivity gap is equally strong as EgoCogNav produces higher elevation of perceived uncertainty during annotated navigation behaviors compared to the baselines.

\noindent\textbf{Ablation study.}
\label{sec:ablation}
Table~\ref{tab:ablation} presents the ablation results. The full model achieves the best trajectory performance across all scenarios and high-uncertainty moments. The largest improvement comes from adding uncertainty prediction alone which reduces FDE by 9.2\% and head error by 8.2\%. This suggests that uncertainty supervision provides useful gradient signal to the shared encoder to encourage representations that are sensitive to decision difficulty and behavior transitions (e.g., scanning, hesitation, and backtracking). Since moments of high perceived uncertainty systematically co-occur with hesitation, direction changes, and head confirming, the encoder learns features that support both cognitive-state estimation and short-horizon forecasting. Neither UCD nor memory alone substantially improves performance, but their combination produces the largest trajectory gains. This complementarity arises because the two modules address different aspects of the forecasting problem. The memory module augments the representation with learned navigation patterns, while UCD modulates decoding according to the predicted uncertainty state. Their combination allows the model to use uncertainty not only as an auxiliary supervised signal, but also as an internal conditioning variable for trajectory and head-motion prediction.

Table~\ref{tab:modality} reports the contribution of each input modality. Motion and goal alone provide limited gains, while adding video substantially improves trajectory and uncertainty prediction. Head history further improves head-motion forecasting, and the full multimodal model achieves the best overall performance. Table~\ref{tab:per_behavior} further breaks results down by annotated behaviors. EgoCogNav achieves the best trajectory performance on \texttt{HES}, \texttt{WRONG}, and \texttt{BACK}, indicating that the method most benefits decision-intensive moments. For head behaviors, improvements are distributed across components where memory alone performs best on \texttt{CONF} and \texttt{LB}, and the full model provides the strongest gain on \texttt{SCAN}.

\begin{table}[t]
\centering
\small
\setlength{\tabcolsep}{4pt}
\renewcommand{\arraystretch}{1.15}
\begin{tabular}{@{}lccc|ccc|ccc@{}}
\toprule
& & & &
\multicolumn{3}{c|}{All Scenarios} &
\multicolumn{3}{c}{High-U Scenarios} \\
\cmidrule(lr){5-7}\cmidrule(lr){8-10}
\textbf{Method} & U & UCD & Mem &
ADE $\downarrow$ & FDE $\downarrow$ & Head $\downarrow$ &
ADE $\downarrow$ & FDE $\downarrow$ & Head $\downarrow$ \\
\midrule
Base module
& -- & -- & -- & .1168 & .2443 & .0785 & .1286 & .2630 & .0790 \\
+\,U prediction
 & \checkmark & -- & -- & .1121 & .2217 & .0721 & .1212 & .2401 & .0722 \\
 +\,UCD
 & \checkmark & \checkmark & -- & .1096 & .2188 & .0712 & .1202 & .2381 & .0716 \\
 +\,Memory
 & \checkmark & -- & \checkmark & .1114 & .2193 & .0707 & .1228 & .2384 & .0710 \\
\textbf{EgoCogNav}
& \checkmark & \checkmark & \checkmark & \textbf{.1051} & \textbf{.2074} & \textbf{.0698} & \textbf{.1155} &
\textbf{.2256} & \textbf{.0704} \\
\bottomrule
\end{tabular}
\caption{\textbf{Ablation study.} Each design choice contributes and UCD and memory show complementary gains when combined.}
\label{tab:ablation}
\end{table}

\begin{table}[t]
\centering
\footnotesize
\setlength{\tabcolsep}{3pt}
\begin{tabular}{lccccc}
\toprule
Input & ADE$\downarrow$ & FDE$\downarrow$ & Head L1$\downarrow$ & $\rho_U\uparrow$ & MAE$_U\downarrow$ \\
\midrule
Motion Only  & 0.156 & 0.314 & 0.0794 & 0.313 & 0.178 \\
Motion+Goal  & 0.155 & 0.313 & 0.0791 & 0.314 & 0.181 \\
+Video       & 0.112 & 0.213 & 0.0713 & 0.778 & 0.0968 \\
+Head        & 0.115 & 0.217 & 0.0708 & 0.780 & 0.0971 \\
Full (+Gaze)         &  \textbf{0.105}   &  \textbf{0.207}   &  \textbf{0.0698}   & \textbf{0.788}   & \textbf{0.0986}   \\
\bottomrule
\end{tabular}
% \vspace{-1mm}
\caption{\textbf{Modality ablation.} Contribution of each input modality to trajectory, head-motion, and uncertainty prediction.}
\label{tab:modality}
\end{table}

\begin{table*}[t]
\centering
\small
\setlength{\tabcolsep}{4pt}
\renewcommand{\arraystretch}{1.15}
\begin{tabular}{@{}l|cc|cc|cc|c|c|c@{}}
\toprule
    &
\multicolumn{6}{c|}{Trajectory Behaviors} & \multicolumn{3}{c}{Head Behaviors} \\
\cmidrule(lr){2-7}\cmidrule(lr){8-10}
    &
\multicolumn{2}{c|}{\texttt{HES}} & \multicolumn{2}{c|}{\texttt{WRONG}} & \multicolumn{2}{c|}{\texttt{BACK}}
 & \texttt{SCAN} & \texttt{CONF} & \texttt{LB} \\
\cmidrule(lr){2-3}\cmidrule(lr){4-5}\cmidrule(lr){6-7}\cmidrule(lr){8-8}\cmidrule(lr){9-9}\cmidrule(lr){10-10}
\textbf{Method} & ADE$\downarrow$ & FDE$\downarrow$ & ADE$\downarrow$ & FDE$\downarrow$ & ADE$\downarrow$ &
FDE$\downarrow$ & Head$\downarrow$ & Head$\downarrow$ & Head$\downarrow$ \\
\midrule
Base module
      & .1192 & .2442 & .0929 & .2013 & .1178 & .2471
      & .0991 & .0820 & .1278 \\
+\,U prediction
      & .1136 & .2214 & .0952 & .1938 & .1082 & .2205
      & .0877 & .0768 & \textbf{.1244} \\
 +\,UCD
      & .1160 & .2232 & .0945 & .1893 & .1092 & .2218
      & .0872 & .0789 & .1248 \\
 +\,Memory
      & .1134 & .2188 & .0946 & .1909 & .1064 & .2169
      & .0869 & \textbf{.0749} & \textbf{.1244} \\
\textbf{EgoCogNav}
 & \textbf{.1112} & \textbf{.2115} & \textbf{.0905} & \textbf{.1825} & \textbf{.1040} & \textbf{.2100}
 & \textbf{.0864} & .0751 & .1277 \\
\bottomrule
\end{tabular}
\caption{\textbf{Ablation for navigation behaviors.} Performance during annotated navigation behaviors for trajectories and head movements.}
\label{tab:per_behavior}
\end{table*}

\subsection{Qualitative Evaluation}
\label{sec:qual}
\begin{figure*}[t]
  \centering
  \includegraphics[width=1\linewidth]{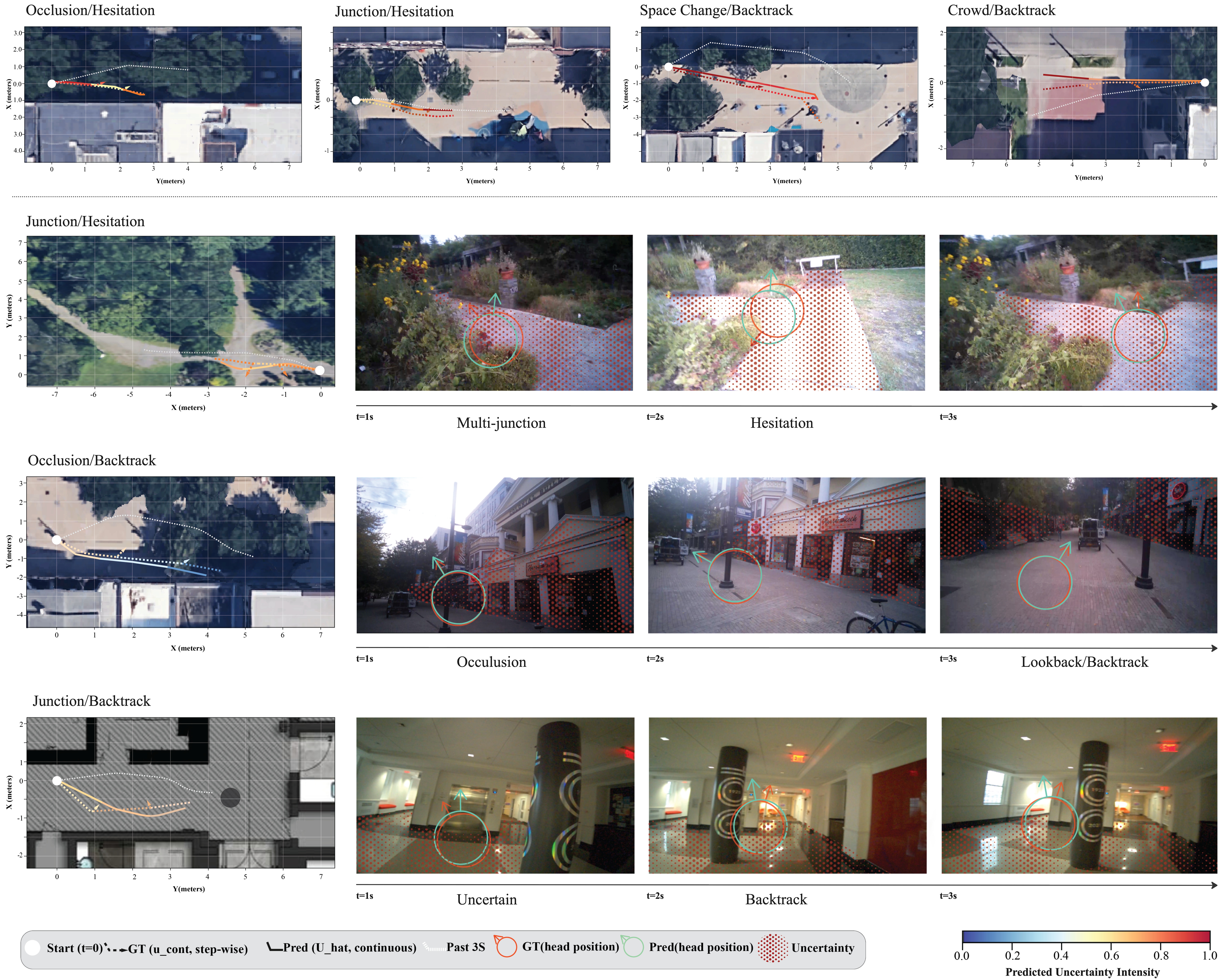}
  \caption{\textbf{Qualitative visualizations.} The top row presents BEV examples under high-uncertainty and behavior-eliciting scenarios. For each scenario on the bottom row, the left panel shows a BEV overlay with past trajectory (gray), ground-truth future, predicted future, and overlaid path uncertainty. The right panels show time-aligned egocentric frames (t+1 to t+3s) with ground-truth (red) and predicted (green) head positions. Environments were also highlighted with red dots for those that trigger uncertain behaviors.}
  \label{fig:bev}
\end{figure*}

We visualize predictions alongside ground truth for trajectories and head motion, and overlay uncertainty as a color-coded intensity along the predicted path shown in Fig.~\ref{fig:bev}. Overall, our model closely tracks ground-truth trajectories and head orientations across conditions while producing uncertainty values that align with decision difficulty environments. In multi-junction scenes, we observe elevated uncertainty before hesitation and scanning. And in occlusion-heavy cases, uncertainty peaks before backtracking. Conversely, in well-specified corridors with clear sight-lines, uncertainty remains low and motion is smooth. These patterns qualitatively support our analysis and indicate that the model captures both route-finding behaviors, and their coupling with perceived uncertainty. From an environment-centric perspective, we observe that scenes rated by participants as more confusing or cognitively demanding can systematically induce higher predicted uncertainty, while visually simple regions correspond to low predicted uncertainty. This alignment between environmental structures and subjective feelings from model outputs suggests that EgoCogNav captures not only route-finding behaviors but also how different environments are experienced.

\noindent\textbf{Failure cases.} We also identify failure cases as illustrated in Fig.~\ref{fig:fail}. In the first scenario, the participant suspected a wrong turn due to heavy occlusion and backtracked to seek an alternative route. Our model instead predicted a return along the same path segment rather than backtracking to other decision points. This indicates insufficient use of long-horizon visual context and episodic scene memory. In the second example, although the predicted path heads in the correct direction, it fails to capture the brief hesitation and look-back behaviors triggered by the changing scene context. These failure cases emphasize the need for stronger global context beyond the immediate scene representations and for explicit multi-hypothesis futures when comparable uncertainty covers multiple plausible routes.

\begin{figure}[t]
  \centering
  \includegraphics[width=1\columnwidth]{figure/failure_h.jpg}
  \caption{\textbf{Failure cases.} Two failure cases highlight limits in long-horizon scene memory.}
  \label{fig:fail}
\end{figure}

\subsection{Limitations and Future Work}
\label{sec:limitation}
There are several limitations noted. First, when salient cues are outside the camera frustum or are heavily occluded, the model operates with partial scene knowledge and degrades performance. Future work can incorporate richer 3D/semantic context to improve disambiguation at complex environments. Second, we predict a single best trajectory and head sequence, and can utilize generative models to sample a distribution of futures can better capture diverse navigation styles under similar cognitive states. Third, this work features uncertainty as main cognitive factor and future work will incorporate additional cognitive/experiential signals such as affect and spatial memory and extend from short-horizon forecasting to hierarchical, longer-horizon planning. Last, CEN is intended as a first-step dataset for cognition-aware egocentric navigation. Future extensions will expand the participant pool, scene diversity, and route types, and incorporate richer annotations such as signage visibility, semantic landmarks, and social dynamics. These additions would support stronger analysis of personalization, environment-level difficulty, and assistive navigation policies.

\vspace{-2mm}
\section{Conclusion}
\label{sec:conclusion}

In this paper, we present several key contributions to the goal of cognition-aware egocentric navigation. First, we introduce the challenging task of jointly forecasting body-centered trajectory, head motion, and perceived path uncertainty from multimodal inputs. This formulation enables the model to reason not only about future motion but also about underlying cognitive states to explain onset behaviors and environmental evaluation. Second, we propose EgoCogNav, an architecture that effectively integrates scene features and sensory cues to accurately forecast human motion. To support this task, we collect the CEN dataset, a 6-hour collection of real-world recordings across 42 diverse environments with synchronized video, gaze, and self-reported uncertainty. Our results indicate that EgoCogNav learns the coupling between perception, perceived uncertainty, and human-like navigation behaviors and improves forecasting performance on held-out navigation recordings. We also discuss the limitations of the project and important future work that is needed toward real-world deployment in social and assistive navigation technologies.

% ---- Bibliography ----
\bibliographystyle{splncs04}
\bibliography{main}

@String(ECCV= {Eur. Conf. Comput. Vis.})

@String(ECCV  = {ECCV})

@inproceedings{damen2018scaling,
  title={Scaling egocentric vision: The epic-kitchens dataset},
  author={Damen, Dima and Doughty, Hazel and Farinella, Giovanni Maria and Fidler, Sanja and Furnari, Antonino and Kazakos, Evangelos and Moltisanti, Davide and Munro, Jonathan and Perrett, Toby and Price, Will and others},
  booktitle={Proceedings of the European conference on computer vision (ECCV)},
  pages={720--736},
  year={2018}
}

@inproceedings{grauman2022ego4d,
  title={Ego4d: Around the world in 3,000 hours of egocentric video},
  author={Grauman, Kristen and Westbury, Andrew and Byrne, Eugene and Chavis, Zachary and Furnari, Antonino and Girdhar, Rohit and Hamburger, Jackson and Jiang, Hao and Liu, Miao and Liu, Xingyu and others},
  booktitle={Proceedings of the IEEE/CVF conference on computer vision and pattern recognition},
  pages={18995--19012},
  year={2022}
}

@inproceedings{li2021ego,
  title={Ego-exo: Transferring visual representations from third-person to first-person videos},
  author={Li, Yanghao and Nagarajan, Tushar and Xiong, Bo and Grauman, Kristen},
  booktitle={Proceedings of the IEEE/CVF Conference on Computer Vision and Pattern Recognition},
  pages={6943--6953},
  year={2021}
}

@inproceedings{wang2023scene,
  title={Scene-aware egocentric 3d human pose estimation},
  author={Wang, Jian and Luvizon, Diogo and Xu, Weipeng and Liu, Lingjie and Sarkar, Kripasindhu and Theobalt, Christian},
  booktitle={Proceedings of the IEEE/CVF Conference on Computer Vision and Pattern Recognition},
  pages={13031--13040},
  year={2023}
}

@article{lv2024aria,
  title={Aria everyday activities dataset},
  author={Lv, Zhaoyang and Charron, Nicholas and Moulon, Pierre and Gamino, Alexander and Peng, Cheng and Sweeney, Chris and Miller, Edward and Tang, Huixuan and Meissner, Jeff and Dong, Jing and others},
  journal={arXiv preprint arXiv:2402.13349},
  year={2024}
}

@article{li2021eye,
  title={In the eye of the beholder: Gaze and actions in first person video},
  author={Li, Yin and Liu, Miao and Rehg, James M},
  journal={IEEE transactions on pattern analysis and machine intelligence},
  volume={45},
  number={6},
  pages={6731--6747},
  year={2021},
  publisher={IEEE}
}

@inproceedings{pan2025lookout,
  title={LookOut: Real-World Humanoid Egocentric Navigation},
  author={Pan, Boxiao and Harley, Adam W and Engelmann, Francis and Liu, C Karen and Guibas, Leonidas J},
  booktitle={Proceedings of the IEEE/CVF International Conference on Computer Vision},
  pages={24977--24988},
  year={2025}
}

@inproceedings{ma2024nymeria,
  title={Nymeria: A massive collection of multimodal egocentric daily motion in the wild},
  author={Ma, Lingni and Ye, Yuting and Hong, Fangzhou and Guzov, Vladimir and Jiang, Yifeng and Postyeni, Rowan and Pesqueira, Luis and Gamino, Alexander and Baiyya, Vijay and Kim, Hyo Jin and others},
  booktitle={European Conference on Computer Vision},
  pages={445--465},
  year={2024},
  organization={Springer}
}

@inproceedings{banerjee2025hot3d,
  title={Hot3d: Hand and object tracking in 3d from egocentric multi-view videos},
  author={Banerjee, Prithviraj and Shkodrani, Sindi and Moulon, Pierre and Hampali, Shreyas and Han, Shangchen and Zhang, Fan and Zhang, Linguang and Fountain, Jade and Miller, Edward and Basol, Selen and others},
  booktitle={Proceedings of the Computer Vision and Pattern Recognition Conference},
  pages={7061--7071},
  year={2025}
}

@inproceedings{li2024egoexo,
  title={Egoexo-fitness: Towards egocentric and exocentric full-body action understanding},
  author={Li, Yuan-Ming and Huang, Wei-Jin and Wang, An-Lan and Zeng, Ling-An and Meng, Jing-Ke and Zheng, Wei-Shi},
  booktitle={European Conference on Computer Vision},
  pages={363--382},
  year={2024},
  organization={Springer}
}

@article{tang2023egotracks,
  title={Egotracks: A long-term egocentric visual object tracking dataset},
  author={Tang, Hao and Liang, Kevin J and Grauman, Kristen and Feiszli, Matt and Wang, Weiyao},
  journal={Advances in Neural Information Processing Systems},
  volume={36},
  pages={75716--75739},
  year={2023}
}

@article{engel2023project,
  title={Project aria: A new tool for egocentric multi-modal ai research},
  author={Engel, Jakob and Somasundaram, Kiran and Goesele, Michael and Sun, Albert and Gamino, Alexander and Turner, Andrew and Talattof, Arjang and Yuan, Arnie and Souti, Bilal and Meredith, Brighid and others},
  journal={arXiv preprint arXiv:2308.13561},
  year={2023}
}

@article{wang2024egonav,
  title={Egonav: Egocentric scene-aware human trajectory prediction},
  author={Wang, Weizhuo and Liu, C Karen and Kennedy III, Monroe},
  journal={arXiv preprint arXiv:2403.19026},
  year={2024}
}

@inproceedings{yang2025path,
  title={PATH-U: A data-driven agent-based wayfinding model incorporating perceived path uncertainty and cognitive strategies in unfamiliar indoor environments},
  author={Yang, Qi and Dubey, Rohit K and Kalantari, Saleh},
  booktitle={Building Simulation},
  volume={18},
  number={2},
  pages={449--471},
  year={2025},
  organization={Springer}
}

@inproceedings{bae2024singulartrajectory,
  title={Singulartrajectory: Universal trajectory predictor using diffusion model},
  author={Bae, Inhwan and Park, Young-Jae and Jeon, Hae-Gon},
  booktitle={Proceedings of the IEEE/CVF Conference on Computer Vision and Pattern Recognition},
  pages={17890--17901},
  year={2024}
}

@inproceedings{gu2022stochastic,
  title={Stochastic trajectory prediction via motion indeterminacy diffusion},
  author={Gu, Tianpei and Chen, Guangyi and Li, Junlong and Lin, Chunze and Rao, Yongming and Zhou, Jie and Lu, Jiwen},
  booktitle={Proceedings of the IEEE/CVF conference on computer vision and pattern recognition},
  pages={17113--17122},
  year={2022}
}

@inproceedings{mao2023leapfrog,
  title={Leapfrog diffusion model for stochastic trajectory prediction},
  author={Mao, Weibo and Xu, Chenxin and Zhu, Qi and Chen, Siheng and Wang, Yanfeng},
  booktitle={Proceedings of the IEEE/CVF conference on computer vision and pattern recognition},
  pages={5517--5526},
  year={2023}
}

@inproceedings{maeda2023fast,
  title={Fast inference and update of probabilistic density estimation on trajectory prediction},
  author={Maeda, Takahiro and Ukita, Norimichi},
  booktitle={Proceedings of the IEEE/CVF international conference on computer vision},
  pages={9795--9805},
  year={2023}
}

@inproceedings{scholler2021flomo,
  title={Flomo: Tractable motion prediction with normalizing flows},
  author={Sch{\"o}ller, Christoph and Knoll, Alois},
  booktitle={2021 IEEE/RSJ International Conference on Intelligent Robots and Systems (IROS)},
  pages={7977--7984},
  year={2021},
  organization={IEEE}
}

@inproceedings{alahi2016social,
  title={Social lstm: Human trajectory prediction in crowded spaces},
  author={Alahi, Alexandre and Goel, Kratarth and Ramanathan, Vignesh and Robicquet, Alexandre and Fei-Fei, Li and Savarese, Silvio},
  booktitle={Proceedings of the IEEE conference on computer vision and pattern recognition},
  pages={961--971},
  year={2016}
}

@article{kothari2021human,
  title={Human trajectory forecasting in crowds: A deep learning perspective},
  author={Kothari, Parth and Kreiss, Sven and Alahi, Alexandre},
  journal={IEEE Transactions on Intelligent Transportation Systems},
  volume={23},
  number={7},
  pages={7386--7400},
  year={2021},
  publisher={IEEE}
}

@article{saadatnejad2023social,
  title={Social-transmotion: Promptable human trajectory prediction},
  author={Saadatnejad, Saeed and Gao, Yang and Messaoud, Kaouther and Alahi, Alexandre},
  journal={arXiv preprint arXiv:2312.16168},
  year={2023}
}

@inproceedings{li2023ego,
  title={Ego-body pose estimation via ego-head pose estimation},
  author={Li, Jiaman and Liu, Karen and Wu, Jiajun},
  booktitle={Proceedings of the IEEE/CVF Conference on Computer Vision and Pattern Recognition},
  pages={17142--17151},
  year={2023}
}

@inproceedings{tome2019xr,
  title={xr-egopose: Egocentric 3d human pose from an hmd camera},
  author={Tome, Denis and Peluse, Patrick and Agapito, Lourdes and Badino, Hernan},
  booktitle={Proceedings of the IEEE/CVF International Conference on Computer Vision},
  pages={7728--7738},
  year={2019}
}

@inproceedings{yi2025estimating,
  title={Estimating body and hand motion in an ego-sensed world},
  author={Yi, Brent and Ye, Vickie and Zheng, Maya and Li, Yunqi and M{\"u}ller, Lea and Pavlakos, Georgios and Ma, Yi and Malik, Jitendra and Kanazawa, Angjoo},
  booktitle={Proceedings of the Computer Vision and Pattern Recognition Conference},
  pages={7072--7084},
  year={2025}
}

@inproceedings{akada20243d,
  title={3d human pose perception from egocentric stereo videos},
  author={Akada, Hiroyasu and Wang, Jian and Golyanik, Vladislav and Theobalt, Christian},
  booktitle={Proceedings of the IEEE/CVF Conference on computer vision and pattern recognition},
  pages={767--776},
  year={2024}
}

@article{mavrogiannis2023core,
  title={Core challenges of social robot navigation: A survey},
  author={Mavrogiannis, Christoforos and Baldini, Francesca and Wang, Allan and Zhao, Dapeng and Trautman, Pete and Steinfeld, Aaron and Oh, Jean},
  journal={ACM Transactions on Human-Robot Interaction},
  volume={12},
  number={3},
  pages={1--39},
  year={2023},
  publisher={ACM New York, NY}
}

@article{farr2012wayfinding,
  title={Wayfinding: A simple concept, a complex process},
  author={Farr, Anna Charisse and Kleinschmidt, Tristan and Yarlagadda, Prasad and Mengersen, Kerrie},
  journal={Transport Reviews},
  volume={32},
  number={6},
  pages={715--743},
  year={2012},
  publisher={Taylor \& Francis}
}

@inproceedings{raubal1999formal,
  title={A formal model of the process of wayfinding in built environments},
  author={Raubal, Martin and Worboys, Michael},
  booktitle={International conference on spatial information theory},
  pages={381--399},
  year={1999},
  organization={Springer}
}

@inproceedings{edward2009cognitive,
  title={Cognitive modeling of virtual autonomous intelligent agents integrating human factors},
  author={Edward, Lydie and Lourdeaux, Domitile and Barthes, Jean-Paul},
  booktitle={2009 IEEE/WIC/ACM International Joint Conference on Web Intelligence and Intelligent Agent Technology},
  volume={3},
  pages={353--356},
  year={2009},
  organization={IEEE}
}

@article{bosse2013modelling,
  title={Modelling collective decision making in groups and crowds: Integrating social contagion and interacting emotions, beliefs and intentions},
  author={Bosse, Tibor and Hoogendoorn, Mark and Klein, Michel CA and Treur, Jan and Van Der Wal, C Natalie and Van Wissen, Arlette},
  journal={Autonomous Agents and Multi-Agent Systems},
  volume={27},
  number={1},
  pages={52--84},
  year={2013},
  publisher={Springer}
}

@book{pan2006computational,
  title={Computational modeling of human and social behaviors for emergency egress analysis},
  author={Pan, Xiaoshan},
  year={2006},
  publisher={Stanford University}
}

@article{raubal2001human,
  title={Human wayfinding in unfamiliar buildings: a simulation with a cognizing agent},
  author={Raubal, Martin},
  journal={Cognitive Processing},
  volume={2},
  number={3},
  pages={363--388},
  year={2001}
}

@article{zhu2023behavioral,
  title={Behavioral, data-driven, agent-based evacuation simulation for building safety design using machine learning and discrete choice models},
  author={Zhu, Runhe and Becerik-Gerber, Burcin and Lin, Jing and Li, Nan},
  journal={Advanced Engineering Informatics},
  volume={55},
  pages={101827},
  year={2023},
  publisher={Elsevier}
}

@article{xie2022simulation,
  title={Simulation of spontaneous leader--follower behaviour in crowd evacuation},
  author={Xie, Wei and Lee, Eric Wai Ming and Lee, Yiu Yin},
  journal={Automation in Construction},
  volume={134},
  pages={104100},
  year={2022},
  publisher={Elsevier}
}

@article{huang2023modeling,
  title={Modeling pedestrian tactical and operational decisions under risk and uncertainty: A two-layer model framework},
  author={Huang, Rong and Zhao, Xuan and Yuan, Yufei and Yu, Qiang and Liu, Chengqing and Daamen, Winnie},
  journal={IEEE Transactions on Intelligent Transportation Systems},
  volume={24},
  number={5},
  pages={5259--5281},
  year={2023},
  publisher={IEEE}
}

@article{hirsh2012psychological,
  title={Psychological entropy: a framework for understanding uncertainty-related anxiety.},
  author={Hirsh, Jacob B and Mar, Raymond A and Peterson, Jordan B},
  journal={Psychological review},
  volume={119},
  number={2},
  pages={304},
  year={2012},
  publisher={American Psychological Association}
}

@article{devlin2014wayfinding,
  title={Wayfinding in healthcare facilities: Contributions from environmental psychology},
  author={Devlin, Ann Sloan},
  journal={Behavioral sciences},
  volume={4},
  number={4},
  pages={423--436},
  year={2014},
  publisher={MDPI}
}

@article{pouyan2024elderly,
  title={The elderly wayfinding performance in an informative healthcare design indoors},
  author={Pouyan, Amir Ehsan and Ghanbaran, Abdul Hamid and Hosseinzadeh, Abbas and Shakibamanesh, Amir},
  journal={Journal of Building Engineering},
  volume={87},
  pages={108843},
  year={2024},
  publisher={Elsevier}
}

@article{onkhar2024evaluating,
  title={Evaluating the Tobii Pro Glasses 2 and 3 in static and dynamic conditions},
  author={Onkhar, V and Dodou, D and De Winter, JCF},
  journal={Behavior Research Methods},
  volume={56},
  number={5},
  pages={4221--4238},
  year={2024},
  publisher={Springer}
}

@article{schafer2011savitzky,
  title={What is a savitzky-golay filter?[lecture notes]},
  author={Schafer, Ronald W},
  journal={IEEE Signal processing magazine},
  volume={28},
  number={4},
  pages={111--117},
  year={2011},
  publisher={IEEE}
}

@inproceedings{escobar2025egocast,
  title={Egocast: Forecasting egocentric human pose in the wild},
  author={Escobar, Maria and Puentes, Juanita and Forigua, Cristhian and Pont-Tuset, Jordi and Maninis, Kevis-Kokitsi and Arbelaez, Pablo},
  booktitle={2025 IEEE/CVF Winter Conference on Applications of Computer Vision (WACV)},
  pages={5831--5841},
  year={2025},
  organization={IEEE}
}

@article{salzmann2023robots,
  title={Robots that can see: Leveraging human pose for trajectory prediction},
  author={Salzmann, Tim and Chiang, Hao-Tien Lewis and Ryll, Markus and Sadigh, Dorsa and Parada, Carolina and Bewley, Alex},
  journal={IEEE Robotics and Automation Letters},
  volume={8},
  number={11},
  pages={7090--7097},
  year={2023},
  publisher={IEEE}
}

@inproceedings{jain2020discrete,
  title={Discrete residual flow for probabilistic pedestrian behavior prediction},
  author={Jain, Ajay and Casas, Sergio and Liao, Renjie and Xiong, Yuwen and Feng, Song and Segal, Sean and Urtasun, Raquel},
  booktitle={Conference on Robot Learning},
  pages={407--419},
  year={2020},
  organization={PMLR}
}

@inproceedings{vellenga2024evaluation,
  title={Evaluation of Video Masked Autoencoders' Performance and Uncertainty Estimations for Driver Action and Intention Recognition},
  author={Vellenga, Koen and Steinhauer, H Joe and Falkman, G{\"o}ran and Bj{\"o}rklund, Tomas},
  booktitle={Proceedings of the IEEE/CVF Winter Conference on Applications of Computer Vision},
  pages={7429--7437},
  year={2024}
}

@article{mavrogiannis2022social,
  title={Social momentum: Design and evaluation of a framework for socially competent robot navigation},
  author={Mavrogiannis, Christoforos and Alves-Oliveira, Patr{\'\i}cia and Thomason, Wil and Knepper, Ross A},
  journal={ACM Transactions on Human-Robot Interaction (THRI)},
  volume={11},
  number={2},
  pages={1--37},
  year={2022},
  publisher={ACM New York, NY}
}

@inproceedings{walker2022influencing,
  title={Influencing behavioral attributions to robot motion during task execution},
  author={Walker, Nick and Mavrogiannis, Christoforos and Srinivasa, Siddhartha and Cakmak, Maya},
  booktitle={Conference on Robot Learning},
  pages={169--179},
  year={2022},
  organization={PMLR}
}

@inproceedings{jeong2025multi,
  title={Multi-modal Knowledge Distillation-based Human Trajectory Forecasting},
  author={Jeong, Jaewoo and Lee, Seohee and Park, Daehee and Lee, Giwon and Yoon, Kuk-Jin},
  booktitle={Proceedings of the Computer Vision and Pattern Recognition Conference},
  pages={24222--24233},
  year={2025}
}

@article{kretzschmar2016socially,
  title={Socially compliant mobile robot navigation via inverse reinforcement learning},
  author={Kretzschmar, Henrik and Spies, Markus and Sprunk, Christoph and Burgard, Wolfram},
  journal={The International Journal of Robotics Research},
  volume={35},
  number={11},
  pages={1289--1307},
  year={2016},
  publisher={SAGE Publications Sage UK: London, England}
}

@article{hirose2023sacson,
  title={Sacson: Scalable autonomous control for social navigation},
  author={Hirose, Noriaki and Shah, Dhruv and Sridhar, Ajay and Levine, Sergey},
  journal={IEEE Robotics and Automation Letters},
  volume={9},
  number={1},
  pages={49--56},
  year={2023},
  publisher={IEEE}
}

@inproceedings{nguyen2023toward,
  title={Toward human-like social robot navigation: A large-scale, multi-modal, social human navigation dataset},
  author={Nguyen, Duc M and Nazeri, Mohammad and Payandeh, Amirreza and Datar, Aniket and Xiao, Xuesu},
  booktitle={2023 IEEE/RSJ International Conference on Intelligent Robots and Systems (IROS)},
  pages={7442--7447},
  year={2023},
  organization={IEEE}
}

@article{song2024vlm,
  title={Vlm-social-nav: Socially aware robot navigation through scoring using vision-language models},
  author={Song, Daeun and Liang, Jing and Payandeh, Amirreza and Raj, Amir Hossain and Xiao, Xuesu and Manocha, Dinesh},
  journal={IEEE Robotics and Automation Letters},
  year={2024},
  publisher={IEEE}
}

@article{maruyama2017simulation,
  title={Simulation-based evaluation of ease of wayfinding using digital human and as-is environment models},
  author={Maruyama, Tsubasa and Kanai, Satoshi and Date, Hiroaki and Tada, Mitsunori},
  journal={ISPRS International Journal of Geo-Information},
  volume={6},
  number={9},
  pages={267},
  year={2017},
  publisher={MDPI}
}

@article{zhu2021follow,
  title={Follow people or signs? A novel way-finding method based on experiments and simulation},
  author={Zhu, Yu and Chen, Tao and Ding, Ning and Chraibi, Mohcine and Fan, Wei-Cheng},
  journal={Physica A: Statistical Mechanics and its Applications},
  volume={573},
  pages={125926},
  year={2021},
  publisher={Elsevier}
}

@article{de2023large,
  title={Large scale simulation of pedestrian seismic evacuation including panic behavior},
  author={De Iuliis, Melissa and Battegazzorre, Edoardo and Domaneschi, Marco and Cimellaro, Gian Paolo and Bottino, Andrea Giuseppe},
  journal={Sustainable Cities and Society},
  volume={94},
  pages={104527},
  year={2023},
  publisher={Elsevier}
}

@inproceedings{dubey2019fusion,
  title={Fusion-based wayfinding prediction model for multiple information sources},
  author={Dubey, Rohit K and Sohn, Samuel S and Hoelscher, Christoph and Kapadia, Mubbasir},
  booktitle={2019 22th international conference on information fusion (FUSION)},
  pages={1--8},
  year={2019},
  organization={IEEE}
}

@inproceedings{caron2021emerging,
  title={Emerging properties in self-supervised vision transformers},
  author={Caron, Mathilde and Touvron, Hugo and Misra, Ishan and J{\'e}gou, Herv{\'e} and Mairal, Julien and Bojanowski, Piotr and Joulin, Armand},
  booktitle={Proceedings of the IEEE/CVF international conference on computer vision},
  pages={9650--9660},
  year={2021}
}

@article{oquab2023dinov2,
  title={Dinov2: Learning robust visual features without supervision},
  author={Oquab, Maxime and Darcet, Timoth{\'e}e and Moutakanni, Th{\'e}o and Vo, Huy and Szafraniec, Marc and Khalidov, Vasil and Fernandez, Pierre and Haziza, Daniel and Massa, Francisco and El-Nouby, Alaaeldin and others},
  journal={arXiv preprint arXiv:2304.07193},
  year={2023}
}

@inproceedings{amit2020discount,
  title={Discount factor as a regularizer in reinforcement learning},
  author={Amit, Ron and Meir, Ron and Ciosek, Kamil},
  booktitle={International conference on machine learning},
  pages={269--278},
  year={2020},
  organization={PMLR}
}

@inproceedings{zhou2019continuity,
  title={On the continuity of rotation representations in neural networks},
  author={Zhou, Yi and Barnes, Connelly and Lu, Jingwan and Yang, Jimei and Li, Hao},
  booktitle={Proceedings of the IEEE/CVF conference on computer vision and pattern recognition},
  pages={5745--5753},
  year={2019}
}

@article{loshchilov2017decoupled,
  title={Decoupled weight decay regularization},
  author={Loshchilov, Ilya and Hutter, Frank},
  journal={arXiv preprint arXiv:1711.05101},
  year={2017}
}

@article{hauke2011comparison,
  title={Comparison of values of Pearson's and Spearman's correlation coefficients on the same sets of data},
  author={Hauke, Jan and Kossowski, Tomasz},
  journal={Quaestiones geographicae},
  volume={30},
  number={2},
  pages={87--93},
  year={2011}
}

@article{qiu2022egocentric,
  title={Egocentric human trajectory forecasting with a wearable camera and multi-modal fusion},
  author={Qiu, Jianing and Chen, Lipeng and Gu, Xiao and Lo, Frank P-W and Tsai, Ya-Yen and Sun, Jiankai and Liu, Jiaqi and Lo, Benny},
  journal={IEEE Robotics and Automation Letters},
  volume={7},
  number={4},
  pages={8799--8806},
  year={2022},
  publisher={IEEE}
}

@inproceedings{kaya2016importance,
  title={Importance of hospital way-finding system on patient satisfaction},
  author={Kaya, S Didem and Ileri, Y Yalcin and Yuceler, Aydan},
  booktitle={Business Challenges in the Changing Economic Landscape-Vol. 2: Proceedings of the 14th Eurasia Business and Economics Society Conference},
  pages={33--40},
  year={2016},
  organization={Springer}
}

@article{mackett2021mental,
  title={Mental health and wayfinding},
  author={Mackett, Roger L},
  journal={Transportation research part F: traffic psychology and behaviour},
  volume={81},
  pages={342--354},
  year={2021},
  publisher={Elsevier}
}

@inproceedings{peebles2023scalable,
  title={Scalable diffusion models with transformers},
  author={Peebles, William and Xie, Saining},
  booktitle={Proceedings of the IEEE/CVF International Conference on Computer Vision},
  pages={4195--4205},
  year={2023}
}

@article{shi2024mtr++,
  title={{MTR++}: Multi-agent motion prediction with symmetric scene modeling and pair-wise interaction},
  author={Shi, Shaoshuai and Jiang, Li and Dai, Dengxin and Schiele, Bernt},
  journal={IEEE Transactions on Pattern Analysis and Machine Intelligence},
  volume={46},
  number={12},
  pages={9787--9803},
  year={2024}
}

@inproceedings{xu2022remember,
  title={Remember intentions: Retrospective-memory-based trajectory prediction},
  author={Xu, Chenxin and Mao, Weibo and Zhang, Wenjun and Chen, Siheng},
  booktitle={Proceedings of the IEEE/CVF Conference on Computer Vision and Pattern Recognition},
  pages={6488--6497},
  year={2022}
}

@article{yu2024fmtp,
  title={{FMTP}: Fine-grained multi-modal trajectory prediction},
  author={Yu, Jiangang and Zhu, Jiahui},
  journal={arXiv preprint arXiv:2407.02558},
  year={2024}
}

@inproceedings{zhang2023remodiffuse,
  title={{ReMoDiffuse}: Retrieval-augmented motion diffusion model},
  author={Zhang, Mingyuan and Guo, Xinying and Pan, Liang and Cai, Zhongang and Hong, Fangzhou and Li, Huirong and Yang, Lei and Liu, Ziwei},
  booktitle={Proceedings of the IEEE/CVF International Conference on Computer Vision},
  pages={364--373},
  year={2023}
}

@article{elfwing2018sigmoid,
  title={Sigmoid-weighted linear units for neural network function approximation in reinforcement learning},
  author={Elfwing, Stefan and Uchibe, Eiji and Doya, Kenji},
  journal={Neural networks},
  volume={107},
  pages={3--11},
  year={2018},
  publisher={Elsevier}
}

@article{keller2020uncertainty,
  title={Uncertainty promotes information-seeking actions, but what information?},
  author={Keller, Ashlynn M and Taylor, Holly A and Bruny{\'e}, Tad T},
  journal={Cognitive Research: Principles and Implications},
  volume={5},
  number={1},
  pages={42},
  year={2020},
  publisher={Springer}
}

@article{brunye2018spatial,
  title={Spatial decision dynamics during wayfinding: Intersections prompt the decision-making process},
  author={Bruny{\'e}, Tad T and Gardony, Aaron L and Holmes, Amanda and Taylor, Holly A},
  journal={Cognitive Research: Principles and Implications},
  volume={3},
  number={1},
  pages={13},
  year={2018},
  publisher={Springer}
}

\end{document}